\documentclass{article}

\usepackage[numbers,compress]{natbib}
\usepackage[preprint]{neurips_2020}
\usepackage[utf8]{inputenc} % allow utf-8 input
\usepackage[T1]{fontenc}    % use 8-bit T1 fonts
\usepackage{hyperref}       % hyperlinks
\usepackage{url}            % simple URL typesetting
\usepackage{booktabs}       % professional-quality tables
\usepackage{amsfonts}       % blackboard math symbols
\usepackage{nicefrac}       % compact symbols for 1/2, etc.
\usepackage{microtype}      % microtypography

\usepackage{lipsum}
\usepackage{wrapfig}
\usepackage[pdftex]{graphicx} 
\usepackage{amsmath, amsthm, amssymb, amsbsy}

\DeclareMathOperator*{\argmin}{argmin}

\usepackage{soul,color}
\usepackage{graphicx}
\usepackage{subfig}
\usepackage{floatrow}

\newtheorem{theorem}{Theorem}

\newtheorem{proposition}[theorem]{Proposition}

\title{Approximate Inference for Spectral Mixture Kernel }

\author{
  Yohan Jung, Kyungwoo Song, Jinkyoo Park \\
  Department of Industrial $\&$ Systems Engineering\\
  KAIST\\
  Daejeon, Republic of Korea \\
  \texttt{ $\{\text{becre1776,gtshs2,jinkyoo.park}\}$@kaist.ac.kr} \\ 
}

\begin{document}

\maketitle

\begin{abstract}
A spectral mixture (SM) kernel is a flexible kernel used to model any stationary covariance function. Although it is useful in modeling data, the learning of the SM kernel is generally difficult because optimizing a large number of parameters for the SM kernel typically induces an over-fitting, particularly when a gradient-based optimization is used. Also, a longer training time is required. To improve the training, we propose an approximate Bayesian inference for the SM kernel. Specifically, we employ the variational distribution of the spectral points to approximate SM kernel with a random Fourier feature. We optimize the variational parameters by applying a sampling-based variational inference to the derived evidence lower bound (ELBO) estimator constructed from the approximate kernel. To improve the inference, we further propose two additional strategies: (1) a sampling strategy of spectral points to estimate the ELBO estimator reliably and thus its associated gradient, and (2) an approximate natural gradient to accelerate the convergence of the parameters. The proposed inference combined with two strategies accelerates the convergence of the parameters and leads to better optimal parameters.
\end{abstract}

%last name check
\section{Introduction}

In constructing a Gaussian process $(\mathcal{GP})$ model, selecting a proper kernel function is vital because the selected kernel determines the overall structure of the target function by specifying the covariance of $\mathcal{GP}$, a prior for the target function. Inspired by 
Bochner's theorem \citep{bochner1959lectures} in which the spectral density has Fourier duality relationship with the stationary kernel, Wilson et al. \citep{wilson2013gaussian} modeled the spectral density of the kernel using a mixture of the Gaussian distribution and obtained a SM kernel by applying a Fourier transform to the modeled spectral density in an attempt to design a flexible kernel. This SM kernel is flexible enough to approximate any stationary kernel because the mixture of the Gaussian distribution can approximate well any spectral density of a stationary kernel \citep{kostantinos2000gaussian}.

The SM kernel has motivated many researchers to devise more expressive kernels based on the spectral density, i.e., a Fourier duality of the kernel. Ulrich et al. \citep{ulrich2015gp} and Parra et al. \citep{parra2017spectral} proposed a cross-spectral mixture (CSM) kernel and a multi-output spectral mixture (MOSM) for a multi-output $\mathcal{GP}$ by modeling the cross-spectral density between the stochastic processes. Remes et al. \citep{remes2017non} propose a non-stationary spectral kernel by modeling the input-dependent spectral density. 

% Remes et al. \cite{remes2018neural} and Xue et al. \cite{xue2019deep} employ a neural network to model the spectral density of a kernel.

Despite the expressive power of the SM kernel and its variants, employing such kernels for a $\mathcal{GP}$ model is limited because they use numerous hyperparameters for the flexible spectral density modeling, increasing the difficulty of the training. Specifically, the training of many of the kernel hyperparameters is prone to an over-fitting, particularly when gradient-based optimization is employed \citep{warnes1987problems}. Moreover, computing the operations needed to update the hyperparameters requires additional time \citep{rasmussen2004gaussian}, which prevents the SM kernel from being applied to the modeling of large-scale data.

To tackle these issues, we propose an approximate Bayesian inference method for an SM kernel. Based on the intuition that learning of a SM kernel is equivalent to the learning of its spectral density distribution, we find the variational posterior distribution of the spectral density by employing the stochastic gradient variational bayes (SGVB) \citep{kingma2013auto}, which is a sampling-based variational inference. To this end, we employ a variational distribution of the spectral points to approximate the spectral density of the SM kernel using sampled spectral points. To learn variational parameters of the spectral points, we first derive the regularized evidence lower bound (ELBO) estimator computed efficiently by the sampled spectral points. We then optimize the ELBO estimator with respect to the variational and other parameters by employing a sampling-based optimization through a reparameterization trick.

To improve the inference, we propose two additional strategies: (1) a sampling strategy of spectral points used to reliably estimate the ELBO and its associated gradient, (2) a natural gradient that reflects the geometric information of the probability density to accelerate the convergence of the parameters. To validate the proposed methods, we run several experiments on kernel matrix approximation, an ablation study of the approximate inference, and a regression task using for large-scale datasets.

% where ${\tau} = x_{1}-x_{2}$ is difference between two inputs $x_{1},x_{2} \in R^{D}$.
%where $x_{1},x_{2} \in R^{D}$ are the inputs

\section{Preliminaries}

\subsection{Spectral Mixture (SM) Kernel}
We describe how the SM kernel is defined by the spectral density modeling. Bochner's theorem states that stationary kernel $k(x_1-x_2)$ can be defined as the Fourier transform of spectral density $p(S)$ as
\begin{align}
% & k(\tau) = \int e^{2\pi i S^{\mathrm{T}} {\tau} } p(S)  dS 
& k(x_1-x_2) = \int e^{2\pi i S^{\mathrm{T}} {(x_1-x_2)} } p(S)  dS 
\end{align}
for the inputs $x_{1},x_{2} \in R^{D}$. Wilson \citep{wilson2013gaussian} devised a SM kernel by representing the spectral density $p(S)$ using a $Q$ mixture of symmetric Gaussian distribution $p(S) = \sum_{q=1}^{Q} w_{q} \left( \frac{ N(s|\mu_{q},\sigma^{2}_{q}) + N(-s|\mu_{q},\sigma^{2}_{q})  }{2} \right)$ with the mean $\mu_{q} = [ \mu_{(q,1)},..,\mu_{(q,D)} ] \in R^{D} $ and the variance $\sigma^{2}_{q} = [ \sigma^{2}_{(q,1)},..,\sigma^{2}_{(q,D)} ] \in R^{D} $, and by then applying the Fourier Transform to $p(S)$ using Eq. (1). The derived SM kernel is expressed as
\begin{align}
& k_{SM}(x_1-x_2) = \sum_{q=1}^{Q} w_{q} \mathrm{exp} \left( -2{\pi}^2 
\left( \sigma_{q}^{T} (x_1-x_2) \right)^2 \right) \mathrm{cos} \left(2\pi {\mu_{q}}^T (x_1-x_2)  \right)
\end{align}

\subsection{Random Fourier Feature (RFF) }
A random Fourier feature \citep{rahimi2008random} approximates the stationary kernel $k(x_{1}-x_{2})$ by applying a Monte Carlo integration to Eq. (1) with $M$ spectral points $\textbf{\em s} = \{ \textbf{\em s}_{i}\}_{i=1}^{M}$ sampled from $p(S)$ 
\begin{align}
k(x_{1}-x_{2}) & \approx \frac{1}{M} \sum_{i=1}^{M} \cos({2\pi \textbf{\em s}_i}^{T}x_1) \cos({2\pi \textbf{\em s}_i}^{T}x_2) 
+ \sin({2\pi \textbf{\em s}_i}^{T}x_1)\sin({2\pi \textbf{\em s}_i}^{T}x_2)  
\end{align}
If we let the feature map $\phi(x,{\textbf{\em s}}) = \frac{1}{\sqrt{M}} \left[\cos{(2\pi{\textbf{\em s}^{T}_{1}}x)},\sin{(2\pi{\textbf{\em s}^{T}_{1}}x)},..,\cos{(2\pi{\textbf{\em s}^{T}_{M}}x)},\sin{(2\pi{\textbf{\em s}^{T}_{M}}x)} \right] \in R^{1\times2M}$, Eq. (3) can be represented as $k(x_{1}-x_{2}) \approx \phi(x_{1},\textbf{\em s})\phi(x_{2},\textbf{\em s})^{T}$.

\subsection{Sparse Spectrum $\mathcal{GP}$}
%\subsection{Sparse Spectrum Gaussian Process}
For the scalable learning of a large dataset, sparse spectrum $\mathcal{GP}$ \citep{lazaro2010sparse}
employs an approximate kernel obtained by the RFF. Let $\textbf{\em s} = \{ \textbf{\em s}_{i}\}_{i=1}^{M}$ be the spectral points used to approximate the stationary kernel by Eq. (3). In addition, let $f$ be a function, with $\mathcal{GP}$ prior, modeling the relation between $X=\{x_{1},..,x_{N}\}$ and $Y=\{y_{1},..,y_{N}\}$. Then, the prior distribution of $f(X) = [f(x_1),..,f(x_N)]$ using the approximate kernel can be defined as
\begin{align}
p\left(f(X)\right) = N\left( f(X);0,\Phi_{\textbf{\em s}}(X)\Phi_{\textbf{\em s}}(X)^{T} \right)
\end{align}
where $\Phi_{\textbf{\em s}}(X) = \left[ \phi(x_{1},\textbf{\em s});...;\phi(x_{N},\textbf{\em s}) \right] \in R^{N \times 2M}$. If the likelihood is assumed as a Gaussian distribution i.e, $p(Y|f(X))=N \left( Y|f(X),{\sigma}^{2}_{\epsilon} I \right)$, the conditional marginal likelihood of the given spectral points $p(Y|X,\textbf{\em s})$ is computed as $N(Y;0,\Phi_{\textbf{\em s}}(X)\Phi_{\textbf{\em s}}(X)^{T} +{{\sigma}^{2}_{\epsilon}}I)$.

To find the optimal spectral points $\textbf{\em s}$ that explain the data well, $-\log{p(Y|X,\textbf{\em s})}$ is minimized with respect to $\textbf{\em s}$. Evaluating $-\log{p(Y|X,\textbf{\em s})}$ uses the memories $O(NM)$ and takes computation time $O(NM^2)$ for computing the inversion and determinant by inversion lemma. When $M$ is much smaller than $N$, training $\mathcal{GP}$ model using $-\log{p(Y|X,\textbf{\em s})}$ takes less training time because the original $\mathcal{GP}$ model using the exact kernel uses the memories $O(N^{2})$ and takes the computation time $O(N^{3})$ \citep{rasmussen2004gaussian}.

%Natural Gradient is known to be useful in updating the probability density parameter 
% We introduce the natural gradient to be used for our approximate inference.

\subsection{Natural Gradient Optimization (NGO) for the Probability Density Parameter}
It is known that natural gradient can be used in efficiently optimizing the probability density parameter \citep{amari1998natural}. We introduce the natural gradient to be used for our approximate inference. Mathematically, given the loss $\mathcal{L}(\theta)$ parameterized by the parameter $\theta$ of the probability density $p_{\theta}(z)$, for a small $\epsilon>0$, the natural gradient $\Tilde{\nabla}_{\theta} {\mathcal{L}}(\theta)$ can be defined as 
\begin{align}
\Tilde{\nabla}_{\theta} {\mathcal{L}}(\theta) = 
\argmin_{\substack{ \{\Delta\theta ;KL( p_{\theta} \Vert p_{\theta + \Delta\theta} ) = \epsilon  \} }}
\mathcal{L}(\theta + \Delta\theta)
\end{align}
Based on the definition, the natural gradient is consistently defined regardless of the parameterization of the probability density. Because the natural gradient updates robustly the probability density parameters due to its consistent characteristics, it can accelerate the convergence for the inference.

% The natural gradient moves the probability density robustly. This property justifies the fast convergence of the inference through a natural gradient.

% The natural gradient also guarantees the constant learning speed $\epsilon$ for the probability density $p_{\theta}(z)$ in the sense of KL divergence. These properties justify the fast convergence of the inference through a natural gradient.

%\vspace{-.4em}

\section{Approximate Inference for SM kernel}
Based on the intuition that the learning of a SM kernel is equivalent to the learning of its spectral density distribution, in this study, we seek to find the variational posterior distribution of the spectral density by employing the stochastic gradient variational bayes (SGVB) \citep{kingma2013auto}. Specifically, this section discusses 1) how to approximate the spectral density of the SM kernel by spectral points sampled from the variational distribution of the spectral points and how to construct the ELBO estimator using these sampled spectral points, 2) how to effectively sample the spectral points from the variational distribution to robustly compute the ELBO estimator, and 3) how to update the variational parameters by using the approximate natural gradient of the ELBO estimator.

\subsection{Regularized Lower Bound Estimator for Variational Sparse Spectrum Approximation}
First, we assume the variational distribution of spectral points $S=\cup_{q=1}^{Q}\{s_{q,1},..,s_{q,m_{q}}\}$ as $q(S) = \prod_{q=1}^{Q}\prod_{i=1}^{m_q} N(s_{q,i} ;{\mu}_{q},{\sigma^{2}_{q}})$ where $m_q$ is the number of spectral points drawn from the $q$-th spectral density component such that the constructed random kernel can approximate the SM kernel with the hyperparameters $\{w_{q},\mu_{q},\sigma^{2}_{q}\}_{q=1}^{Q}$. If we sample the spectral points $\textit{\textbf{s}}_{q,i}$ from $N(s_{q,i} ;{\mu}_{q},{\sigma^{2}_{q}})$ as $\textit{\textbf{s}}_{q,i}=\mu_{q} + \sigma_{q} \circ \epsilon_i $ with $\epsilon_i \sim N(\epsilon;0,I)$ by the reparameterization trick, we can define the random feature map $\phi_{\mathrm{SM}}(x;\textit{\textbf{s}})$ with the sampled spectral points $\textit{\textbf{s}}=\cup_{q=1}^{Q}\{\textit{\textbf{s}}_{q,1},..,\textit{\textbf{s}}_{q,m_{q}}\}$
\begin{align}
    \phi_{\mathrm{SM}}(x;\textit{\textbf{s}}) = 
    \Big[ \sqrt{w_{1}}  \phi \left(x, \{\textit{\textbf{s}}_{1,i} \}_{i=1}^{m_1}\right), ... ,\sqrt{w_{Q}}  \phi \left(x,\{\textit{\textbf{s}}_{Q,i}\}_{i=1}^{m_Q} \right) \Big] \in R^{1 \times 2M}
\end{align}
where $\phi$ is the defined feature map used to calculate Eq. (3) and $M=\sum_{q=1}^{Q} m_q$. By employing this feature map $\phi_{\mathrm{SM}}(x;\textit{\textbf{s}})$ for the entire dataset $X=\{x_{1},..,x_{N}\}$, we can define the feature matrix as $\Phi^{\mathrm{SM}}_{\textit{\textbf{s}}}(X)=[\phi_{\mathrm{SM}}(x_1;\textit{\textbf{s}});...;\phi_{\mathrm{SM}}(x_N;\textit{\textbf{s}})]\in R^{N\times 2M}$. The feature matrix can then be used to construct the unbiased estimator $\Phi^{\mathrm{SM}}_{\textit{\textbf{s}}}(X)\Phi^{\mathrm{SM}}_{\textit{\textbf{s}}}(X)^{T}$ to satisfy $\mathrm{E}\big[\Phi^{\mathrm{SM}}_{\textit{\textbf{s}}}(X)\Phi^{\mathrm{SM}}_{\textit{\textbf{s}}}(X)^{T} \big] = K_{\mathrm{SM}}(X,X)$. Proposition 1 states the error bound of this estimator.

\begin{proposition}
Let us denote $W_{0}={ \big( \sum_{q=1}^{Q} w^2_{q} \big) }^{1/2}$ and $m_0=\mathrm{min} \{m_1,..,m_Q\}$. Then, for a small $\epsilon >0$, the error bound of $\Phi^{\mathrm{SM}}_{\textit{\textbf{s}}}(X)\Phi^{\mathrm{SM}}_{\textit{\textbf{s}}}(X)^{T}$ using the matrix spectral norm ${\Vert \cdot \Vert}_{2}$ is obtained as
\begin{align*}
\mathrm{Pr} \Big(  {\big \Vert \Phi^{\mathrm{SM}}_{\textbf{s}}(X)\Phi^{\mathrm{SM}}_{\textbf{s}}(X)^{T}  - K_{\mathrm{SM}}(X,X) \big \Vert }_{2}  \geq \epsilon  \Big) 
\leq N \exp{
\Big( \frac{ -3{\epsilon}^2 m_0}{ W_{0}N \big( 6{ \Vert K_{\mathrm{SM}}(X,X) \Vert}_{2} + 4\epsilon  \big) } \Big) }
\end{align*}
\end{proposition}
% + 2WB \epsilon

Using the approximate SM kernel $\Phi^{\mathrm{SM}}_{\textit{\textbf{s}}}(X)\Phi^{\mathrm{SM}}_{\textit{\textbf{s}}}(X)^{T}$ with the equal number of spectral points $m_q=m$ for $q\in\{1,..,Q\}$, we can derive the regularized lower bound estimator $\hat{\mathcal{L}}_{N}$ as
\begin{align}
\log{p(Y|X)} &\geq \int \log{p(Y|X,S)}q(S) d S - KL(q(S)||p(S)) = \mathcal{L}  \nonumber \\
&\approx  \frac{1}{N}\sum_{n=1}^{N} \log {p(Y|X,\textbf{\em s}^{(n)})} - KL(q(S)||p(S)) = \hat{\mathcal{L}}_{N} 
\end{align}
where $\textbf{\em s}^{(n)}$ indicates the $n$-th set of the spectral points sampled from $q(S)$ and $\log {p(Y|X,\textbf{\em s}^{(n)})}$ is evaluated as $\log{N(Y;0,\Phi^{\mathrm{SM}}_{\textbf{\em s}^{(n)}}(X) \Phi^{\mathrm{SM}}_{ \textbf{\em s}^{(n)}}(X)^{T} + {\sigma}^{2}_{\epsilon}I)}$. $p(S)$ is a prior distribution of spectral density expressed as $\prod_{q=1}^{Q}\prod_{i=1}^{m_q} N(s_{q,i} ;\Tilde{{\mu}}_{q,i},{\Tilde{\sigma}^{2}_{q,i}})$ where $\Tilde{{\mu}}_{q,i}$ and ${\Tilde{\sigma}^{2}_{q,i}}$ are initialized based on the prior knowledge. The term $KL\left(q(S)||p(S)\right)$ prevents the model from being over-fitted to the training data. To update the parameters, we evaluate $\hat{\mathcal{L}}_{N}$ and compute its gradient with respect to the variational parameters of the spectral points $\{\mu_{q},\sigma^{2}_{q}\}_{q=1}^{Q}$, weight parameters $\{w_{q}\}_{q=1}^{Q}$, and noise parameter ${\sigma}^{2}_{\epsilon}$.

We cast the problem of learning the SM kernel into the problem of estimating the variational parameters of the spectral points. The described procedure, denoted as SVSS, is our basic approximate inference method. In the following two subsections, we propose strategies to improve the SVSS.

\begin{figure*}[t]
\centering
    {\includegraphics[width=0.85\textwidth]{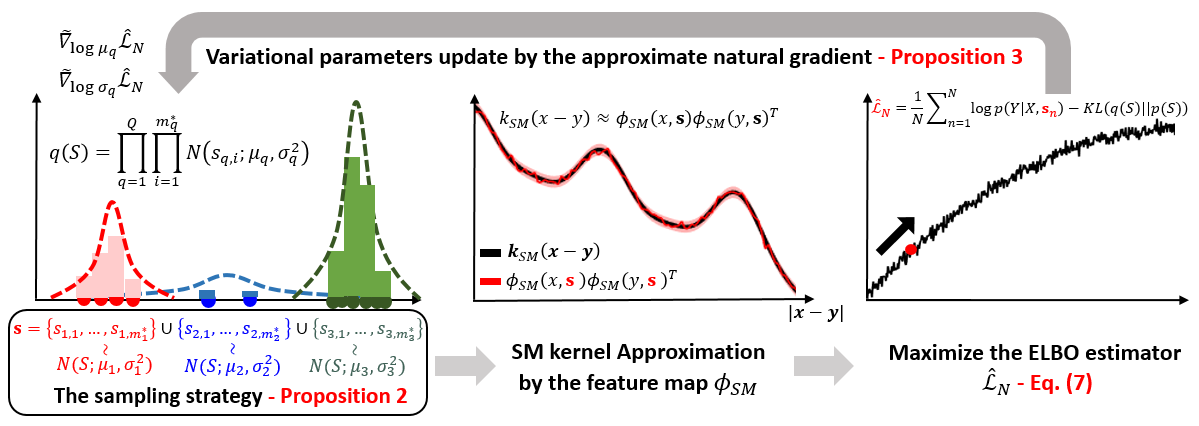}} \hfill
    \caption{This figure describes the outline of the proposed approximate inference for SM kernel.}
\end{figure*}

\subsection{Efficient Sampling Strategy for Spectral Points}

As shown in Eq. (7), the value of $\hat{\mathcal{L}}_{N}$ might fluctuate depending on the sampled spectral point $\textit{\textbf{s}} \sim q(S)$ used to construct $\Phi^{\mathrm{SM}}_{\textit{\textbf{s}}}(X)\Phi^{\mathrm{SM}}_{\textit{\textbf{s}}}(X)^{T}$, especially when the number of total spectral points $M$ is small. Because the large volatility of $\hat{\mathcal{L}}_{N}$ will likely cause instability in learning, we propose a sampling strategy for spectral points to minimize such volatility. Specifically, we find the optimal ratios of the spectral points $\{m_q/\sum_{q=1}^{Q} m_q\}_{q=1}^{Q}$ for the spectral points $\textit{\textbf{s}}$ sampled from $q(S)$ such that the sampled spectral points minimize the sum of the variance for each element, i.e., $\sum_{i,j=1}^{N} \mathrm{Var} \big( [\Phi^{\mathrm{SM}}_{\textit{\textbf{s}}}(X)\Phi^{\mathrm{SM}}_{\textit{\textbf{s}}}(X)^{T}]_{i,j} \big)$. We consider only the upper off-diagonal entries of $\Phi^{\mathrm{SM}}_{\textit{\textbf{s}}}(X)\Phi^{\mathrm{SM}}_{\textit{\textbf{s}}}(X)^{T}$ because this matrix is symmetric and diagonal terms are deterministic. Proposition 2 states the optimization problem obtaining the optimal ratios of the spectral points.
\begin{proposition}
Given the set of inputs $X=\{x_n\}_{n=1}^{N}$, let us define the set of pairwise distances $\{\tau_p\}_{p=1}^{P}$ where $\tau_p=|x_i-x_j|$ for some $i,j \in \{1,..,N\}$ and $i\neq j$. Let $m_{q}$ be the number of spectral points sampled from the variational distribution $N(\mu_q,\sigma^{2}_{q})$, and $M = \sum_{q=1}^{Q} m_{q} $ be the total number of spectral points. For the set of $\{\tau_{p}\}_{p=1}^{P}$, the optimal ratio ${p^{*}_q}=m^{*}_q/M$ to minimize $  \sum_{i<j} \mathrm{Var} \big( [\Phi^{\mathrm{SM}}_{\textbf{s}}(X)\Phi^{\mathrm{SM}}_{\textbf{s}}(X)^{T}]_{i,j} \big)$ is obtained as
\begin{align}
p^{*}_q =  \frac{ w_{q} 
{ \left[ \sum_{p=1}^{P} g_{q}(\tau_{p})   \right]   }^{1/2} } 
{ \sum_{q=1}^{Q} w_{q} {\left[ \sum_{p=1}^{P} g_{q}(\tau_{p}) \right]    }^{1/2} }
\end{align}
where $g_{q}(\tau) = 1 + k_{q}(2\tau) + k^{2}_{q}(\tau)$ and $k_{q}(\tau) = \exp{ \left (  -2{\pi}^2 (\tau^{T} \sigma_{q})^2 \right) } \cos{ \left(   2\pi \mu^{T}_{q}\tau   \right) }$. The integer $m^{*}_{q}$ is determined as $\mathrm{max}\{1,{\lfloor Mp^{*}_q \rceil}\}$ where ${\lfloor m \rceil}$ denotes the integer closest to $m$.
\end{proposition}
Proposition 2 states that the optimal numbers for the sampled spectral points $\textit{\textbf{s}}=\cup_{q=1}^{Q}\{\textit{\textbf{s}}_{q,1},..,\textit{\textbf{s}}_{q,m_{q}}\}$ are determined depending on the variational parameters, weight parameters, and inputs. When we apply this sampling strategy to the SVSS for a large dataset, the total number of the pairwise distances $P=N(N-1)/2$ is too large to be efficiently computed for every iteration. Specifically, we restrict $P$ by selecting a small subset of data. For every iteration, we randomly select the subset $\{\tau_{p_i}\}_{i=1}^{Pr} \subset \{\tau_p\}_{p=1}^{P}$ with the rate $r\in(0,1)$ used for calculating $p^{*}_{q}$. This random sampling will be validated in the experiment.

\subsection{Approximate Natural Gradient Update for Variational Parameters}

Once the ELBO estimator $\hat{\mathcal{L}}_{N}$ is computed, its stochastic gradients with respect to variational parameters of $q(S)$, as well as to other parameters, are computed and used to update the parameters. To accelerate the convergence of the parameters, we propose using the approximate natural gradient easily computed from the original gradient. We consider that variational parameters $\mu_q$ and $\sigma_q$ for $q\in\{1,..,Q\}$ are updated in the logarithm domain because these parameters should remain positive.

\begin{proposition}[Approximate Natural Gradient in the Log Domain]
Let $\mu^{(t)}_q$ and $\sigma^{(t)}_q$ be the $t$-th iterated parameters of $N(\mu_q,\sigma^2_q)$ which is $q$-th component distribution for $q(S)$. The natural gradient of $\hat{\mathcal{L}}_{N}$ w.r.t $\mu_q$ and $\sigma_q$ on log domain, i.e. $\Tilde{\nabla}_{\log{\mu_q}}$ and $\Tilde{\nabla}_{\log{\sigma_q}}$, can be approximated as  
\begin{align}
\Tilde{\nabla}_{\log{\mu_q}} \hat{\mathcal{L}}_{N}   &\approx  \left( \frac{\sigma^{(t+1)}_q} {\mu^{(t)}_q} \right)^{2} \circ {\nabla}_{\log{\mu_q}} \hat{\mathcal{L}}_{N} 
\hspace{3 em}
\Tilde{\nabla}_{\log{\sigma_q}} \hat{\mathcal{L}}_{N}  \approx  \frac{1}{2} \ \nabla_{\log{\sigma_q}} \hat{\mathcal{L}}_{N}
\end{align}
\end{proposition}
under the condition $\Big|  \left( \frac{\sigma^{(t+1)}_q} {\mu^{(t)}_q} \right)^{2} \circ {\nabla}_{\log{\mu_q}} \hat{\mathcal{L}}_{N}  \Big| < 1$ and $ \Big| \nabla_{\log{\sigma_q}} \hat{\mathcal{L}}_{N} \Big| <1$ in element-wise sense. These constraints are satisfied by normalizing the revised gradient by its 2-norm ${\Vert \cdot \Vert}_{2}$. The derived gradients are used to update the parameters using the optimizer with the adaptive learning rate.

\section{Related Work}

To train scalably the parameters of the kernel for large-scale data, variational inducing input method \citep{titsias2009variational,hensman2013gaussian,hensman2015scalable} and sparse spectrum method \citep{lazaro2010sparse,gal2015improving,hoang2017generalized} have been proposed. The variational inducing input (VFE) method introduces a small number of inducing inputs such that the variational distribution of the function-values on the inducing inputs best approximates the prior distribution of the function-values on all inputs. The sparse spectrum method employs the approximate kernel matrix obtained by a random Fourier feature (RFF) \citep{rahimi2008random} to represent a $\mathcal{GP}$ prior. Among the sparse spectrum methods, SSGP \citep{lazaro2010sparse} optimizes the spectral points used for constructing the approximate kernel. To relax the over-fitting issue of SSGP, VSSGP \citep{gal2015improving} applies a variational approximation to the spectral points. VSSGP independently estimates the hyperparameters of the kernel and variational parameters. However, the variational approximation of spectral points can directly approximate the spectral density of kernel through the designed feature map while relaxing the over-fitting. Being different from VSSGP, our approach train the parameters of the SM kernel by inferencing the variational distribution of spectral points. In addition, we consider the sampling strategy and approximate natural gradient to improve the performance of the approximate inference. Because of these differences, our method efficiently train the SM kernel while employing a smaller number of parameters.

% is similar to VSSGP in that both methods employ the variational inference to 
% learn the SM kernel. However, our approach differs from VSSGP in that we approximate the spectral density of the SM kernel and then directly estimate the hyperparameters of the SM kernel. VSSGP does not relate the variational approximation of the spectral points to the spectral density of the SM kernel. 

\section{Experiments}

In the first experiment, we show that our sampling strategy for spectral points helps approximate the SM kernel more accurately. In the second experiment, we validate that the proposed estimator, sampling strategy, and approximate natural gradient improve the inference for the SM kernel through an ablation study. In the last experiment, we conduct a regression task on a large-size UCI dataset \citep{Dua:2019} to evaluate the performance of the proposed approximate inference. We also include the results of additional experiments in the supplementary material.

\subsection{SM Kernel Gram Matrix Approximation}

\vspace{-1.em}
\begin{figure*}[ht]
\centering
    \subfloat[\small{$\{w_{q}\}_{q=1}^{Q}\sim U(0,20)$}]
    {\includegraphics[width=0.48\textwidth]{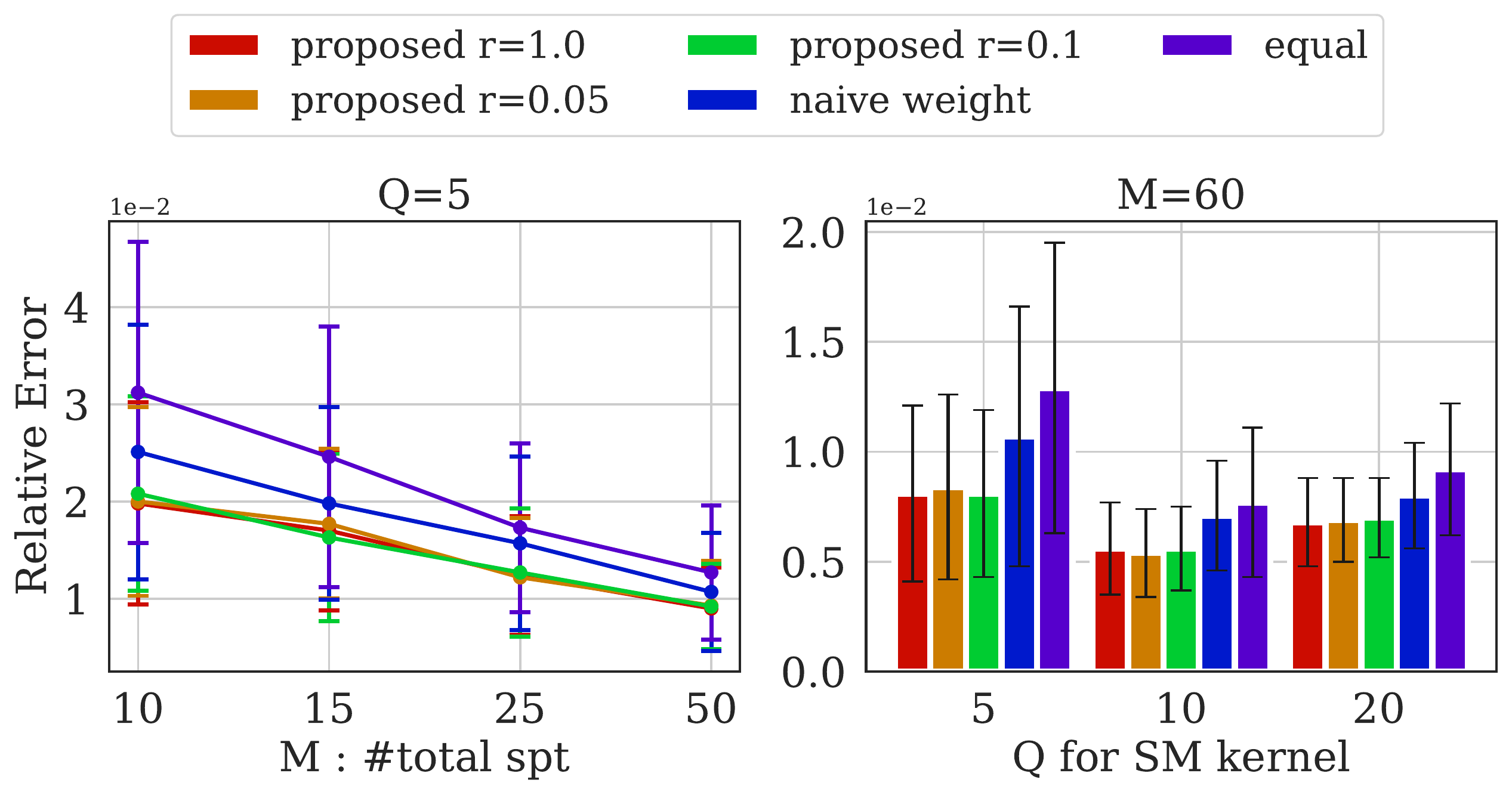}} \hfill 
    \subfloat[\small{$\{w_{q}\}_{q=1}^{Q}\sim U(0.99,1.01)$}]
    {\includegraphics[width=0.48\textwidth]{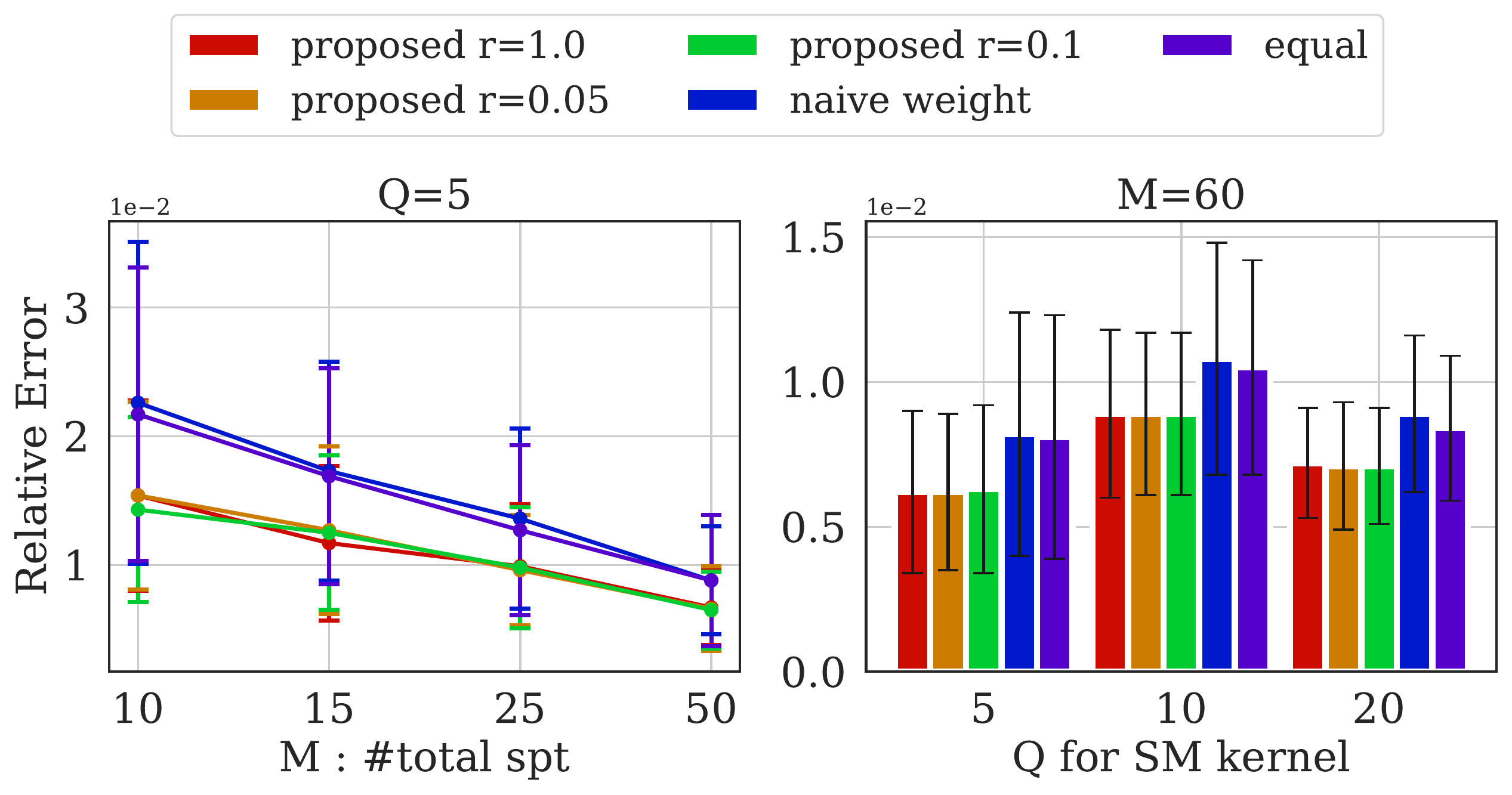}}    
    \caption{SM kernel approximation on synthetic dataset $(N=100 \ D=1)$: (a)  $\{w_{q}\}_{q=1}^{Q}$ randomly initialized by $U(0,20)$ and (b) $\{w_{q}\}_{q=1}^{Q}\sim U(0.99,1.01)$.}   
\end{figure*}

%\vspace{-1.em}

\begin{figure*}[ht]
\centering
    \subfloat[\small{Concrete $(N=1,030 \ D=8)$}]
    {\includegraphics[width=0.48\textwidth]{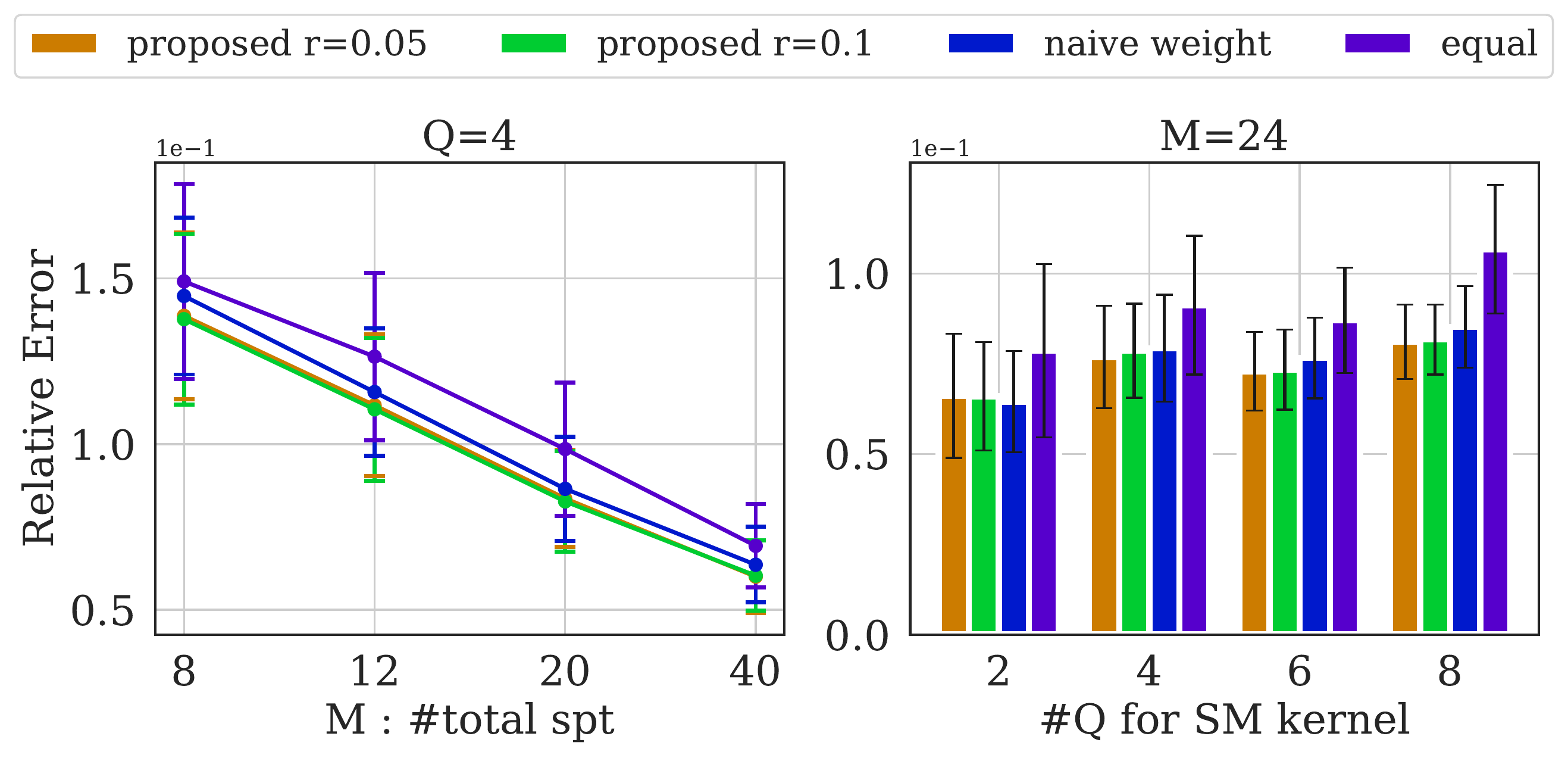}} \hfill
    \subfloat[\small{Parkinsons $(N=5,875 \ D=20)$}]
    {\includegraphics[width=0.48\textwidth]{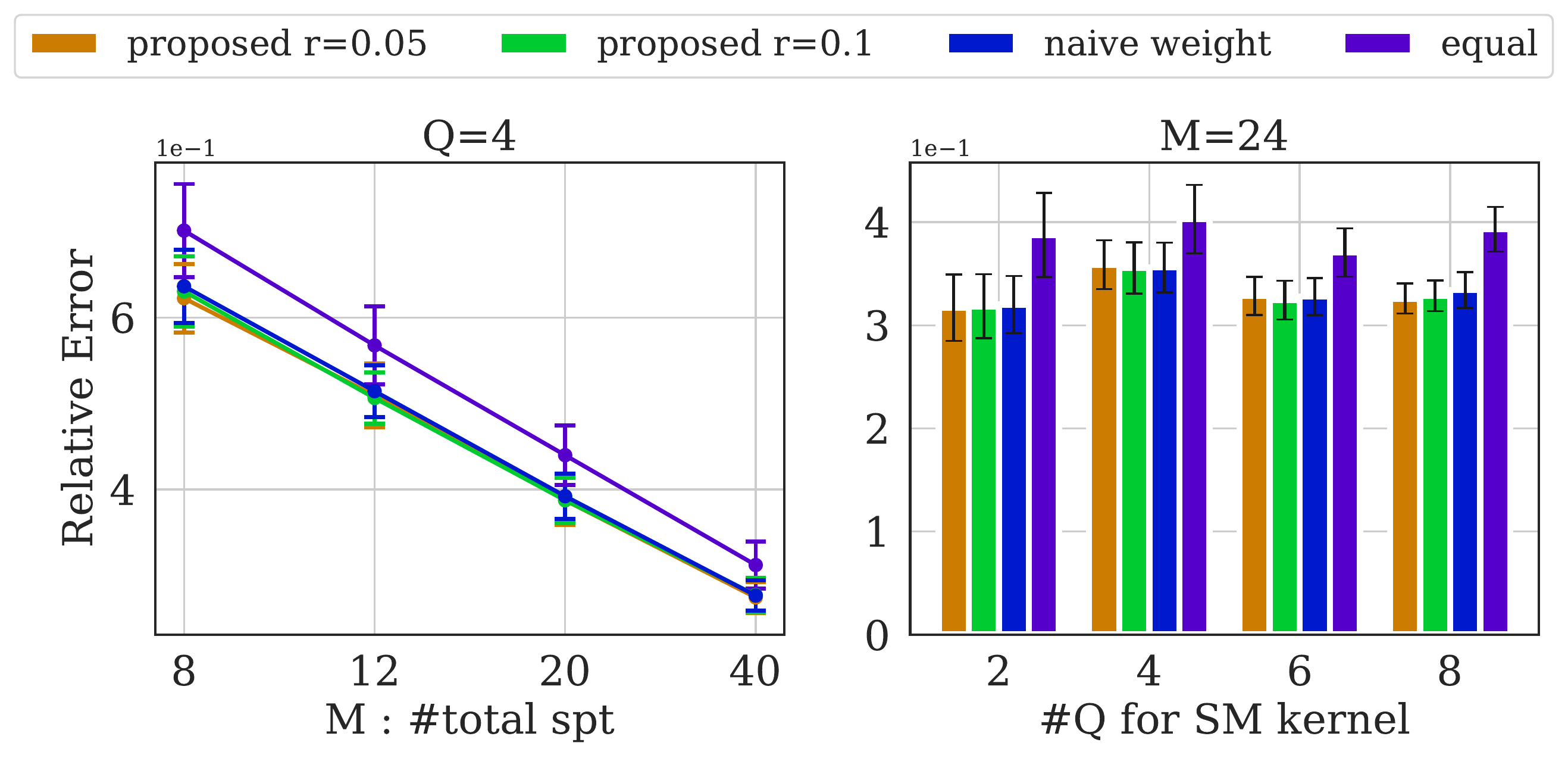}}     
    \caption{SM kernel approximation on (a) Concrete dataset and (b) Parkinsons dataset. The parameters of SM kernel are initialized in the same way as the experiment shown in Figure 2(a).}
\end{figure*}

We verify that the proposed sampling strategy enhances the SM kernel approximation over naive sampling approaches with equal sampling and naive weight sampling of the spectral points using $p_{q}=1/Q$ and $p_{q}=w_q / \sum_{q=1}^{Q} w_q$, respectively. Also, we show that even if a small number of inputs is used to calculate $\{p^{*}_{q}\}_{q=1}^{Q}$ by Eq. (8), our strategy can maintain the quality of the kernel approximation. To validate the sampling's effect in a general setting, we consider the different cases of the number of total spectral points $M$ and the number of mixture components $Q$ for the SM kernel.

% We randomly initialize $\{\mu_{q}\}_{q=1}^{Q}$ by $U(0,5)$ and $\{\sigma_{q}\}_{q=1}^{Q}$ by $U(0,0.1)$.

First, we approximate the small-sized SM kernel $K\in R^{N \times N}$ with $N=100$. We use $X= \{0,.01,..,.99\}$ as the kernel inputs. For the hyperparameters, we initialize the weights of SM kernel in two ways by applying a uniform distribution $U$; $\{w_{q}\}_{q=1}^{Q} \sim U(0,20)$ and $\{w_{q}\}_{q=1}^{Q} \sim U(0.99,1.01)$. We consider these cases to show that the proposed sampling strategy approximates the SM kernel well regardless of its weight parameters. To investigate how the amount of the data used to compute $p^{*}_{q}$ affects the kernel approximation, we use only $r$ fraction among $P=N(N-1)/2$ pair-wise input distances with $r\in\{.05,.1,1\}$. For example, $r=1$ denotes that all data are used for calculating $p^{*}_{q}$. We evaluate the approximation quality based on the relative error ${\Vert K-\hat{K} \Vert}_{F} / {\Vert K \Vert}_{F}$ using the Frobenius norm $F$.

Figures 2(a) and 2(b) describe the results of $\{w_{q}\}_{q=1}^{Q} \sim U(0,20)$ and $\{w_{q}\}_{q=1}^{Q} \sim U(0.99,1.01)$. In Figure 2(a), the proposed method approximates the kernel matrix most accurately, followed by naive weight sampling and equal sampling. By comparing the results of $r=1.0$ and $r=0.05$ in both graphs of (a), we can conclude that using a small fraction of the inputs for calculating $p^{*}_{q}$ does not degrade the performance under various conditions for $M$ and $Q$. In Figure 2(b), similarly, the proposed method outperforms both the naive weight sampling and equal sampling at different values of $M$ and $Q$. We believe that the improved performance is due to the optimal ratio $p^{*}_{q}$ which is computed by the variational parameters $\{\mu_q,\sigma^{2}_{q}\}_{q=1}^{Q}$ and inputs $\{\tau_p\}_{p=1}^{P}$ in addition to $\{w_{q}\}_{q=1}^{Q}$.

% The hyperparameters of the SM kernel are initialized in the same way as in the \hl{Figure 2(a)} experiment.

Additionally, we apply our method on a large-scale high-dimensional real dataset: the Concrete dataset $(N=1,030 \ D=8)$ in Figure 3(a) and the Parkinsons dataset $(N=5,875 \ D=20)$ in Figure 3(b). Because $N$ is large, we consider the restricted value of $P=1,000,000$ and data rate of $r\in\{0.05,0.1\}$. We can see that our sampling strategy is the most effective at approximating the SM kernel in the real dataset.

\begin{figure*}[t]
%\begin{figure*}[h]
\centering
    \subfloat[$Q=4, M \in \{12,20,60\}$]
    {\includegraphics[width=0.95\textwidth]{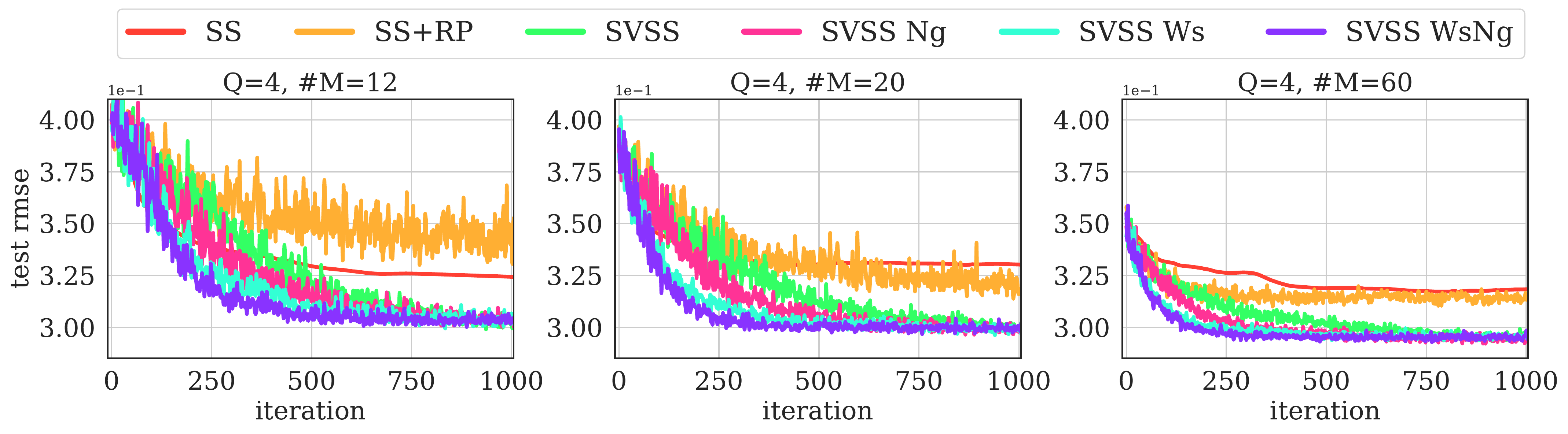}} \vspace{-1em}
    
    \subfloat[\small{$Q=8, M \in \{24,40,120\}$ } ]
    {\includegraphics[width=0.95\textwidth]{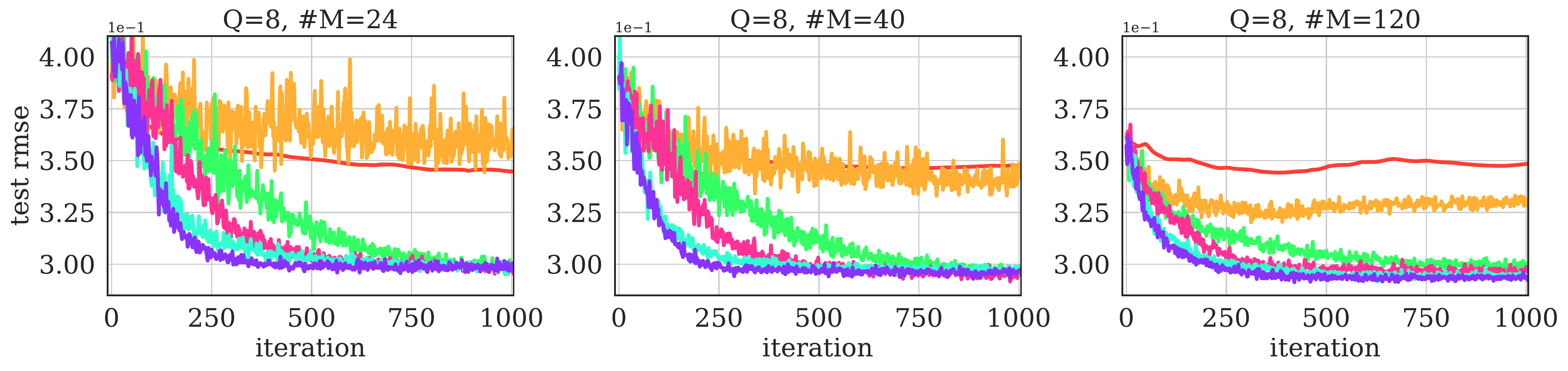}}

\caption{Approximate inference of SM kernel for SkillCraft ($N=3,325 \ D=18$): (a) and (b) each inference method compared for $Q\in\{4,8\}$ and $M\in\{3Q,5Q,15Q\}$, respectively.}
\end{figure*}

%  This approach is similar to the SM kernel inference in A la Carte \cite{yang2015carte} without the fastfood \cite{le2013fastfood} designed for a fast computation of a high dimensional dataset.

\subsection{Approximate Inference for SM kernel }

We investigate the effectiveness of the proposed inference methods composed of the regularized lower bound estimator $\hat{\mathcal{L}}_{N}$, the sampling strategy of proposition 2, and the approximate natural gradient of proposition 3 by conducting an ablation study comparing the following inference methods:

\begin{itemize}
  \item SS: denotes the variation of sparse spectrum $\mathcal{GP}$ \citep{lazaro2010sparse} used to train the SM kernel approximated by Eq. (6). SS optimizes $\{w_{q},\mu_{q},\sigma^{2}_{q}\}_{q=1}^{Q}$ with the given $\{\epsilon_{i}\}_{i=1}^{QM} \sim N(0,I)$ without a reparameterization trick (RP).
  This approach is similar to the SM kernel inference in A la Carte \citep{yang2015carte} without applying the fastfood \citep{le2013fastfood} designed for a fast computation of a high dimensional dataset.
  \vspace{-.25em}
  \item SVSS: optimizes the derived ELBO estimator Eq. (7) using the SGVB. We validate the $\hat{\mathcal{L}}_{N}$ with $N=1$. This is the basic inference method proposed in this study.
  \vspace{-.25em}
 \item SVSS-Ws: denotes the SVSS combined with the weight sampling strategy of Proposition 2.
  \vspace{-.25em}
  \item SVSS-Ng: denotes the SVSS combined with the natural gradient of Proposition 3. 
  \vspace{-.25em} 
  \item SVSS-WsNg: denotes the SVSS combined with the weight sampling strategy of Proposition 2 and the natural gradient of Proposition 3.
  \vspace{-.25em}
\end{itemize}

% After 5 repetitive experiments, the statistical results are obtained. 

% We use the UCI datasets in \cite{Dua:2019}: Skillcrafts, Parkinsons, and Elevators. Each data set is randomly separated into a 90$\%$ training set and 10$\%$ test set. After 5 repetitive experiments, we report the averaged value of the root mean square error (RMSE) of the prediction on the test data during training.

We use the UCI datasets in \citep{Dua:2019}: Skillcraft, Parkinsons, and Elevators. We run 5 repetitive experiments and obtain the statistical result. For each experiment, each data set is randomly separated into a 90$\%$ training set and 10$\%$ test set. In this experiment section, we report only the averaged value of the root mean square error (RMSE) of the prediction on the test data during training.

%\{4,6,8\}$
%%%%%%%%%%%%%%%%%%%%%%%%%%%%%%%%%%%%%%%%%%%
% column 4 figures
%%%%%%%%%%%%%%%%%%%%%%%%%%%%%%%%%%%%%%%%%%%
Figure 4 describes the prediction results of the Skillcraft dataset with $Q\in \{4,8\}$ and $M\in \{3Q,5Q,15Q\}$ using the approximate kernel $\Phi^{\mathrm{SM}}_{\textbf{s}}(X)\Phi^{\mathrm{SM}}_{\textbf{s}}(X)^{T}$. By comparing SVSS with the SS and the SS with the reparameterization trick (SS + RP) for all cases, we can confirm that the regularizer KL term in $\hat{\mathcal{L}}_{N=1}$ helps relax the over-fitting and find better parameters. Notably, its effectiveness becomes clear as $Q$ increases, which implies that the SVSS approach is beneficial for learning numerous parameters. By comparing the results between SVSS, SVSS-Ng, SVSS-Ws, and SVSS-WsNg, we can see that the proposed sampling strategy and natural gradient accelerates the convergence of the variational parameters in all cases of $Q$ and $M$. Especially, the use of a sampling strategy and natural gradient creates the synergy to accelerate the parameter learning. Similar results are obtained for the other datasets. These results are provided in the supplementary material.

\subsection{UCI Dataset Regression Task}

To evaluate the approximate inference quality, we conduct a regression task specifically for the large-scale datasets that are difficult to be trained using a conventional inference method. For the comparison, we consider the following approximation methods in addition to the baseline inference methods compared in the previous experiments: 
\begin{itemize}
  \item  VFE: denotes the variational inducing input method \citep{titsias2009variational}, which is a representative scalable inference method in $\mathcal{GP}$. We use not only the SM kernel but also the RBF (ARD) kernel because the RBF is the one of the most widely used kernels in the $\mathcal{GP}$ model. 
  \vspace{-0.3em}
  \item VSS: denotes the variatioanl sparse spectrum approximation of $\mathcal{GP}$ \citep{gal2015improving} which improves the sparse spectrum $\mathcal{GP}$ \citep{lazaro2010sparse} by variational approximation of the spectral points and the inputs of the data through the VI \citep{jordan1999introduction,hoffman2013stochastic}. This method assumes that the SM kernel is used.
  \vspace{-0.3em}  
\end{itemize}

\begin{figure*}[t]
%\begin{figure*}[h]
\centering

    {\includegraphics[width=0.95\textwidth]{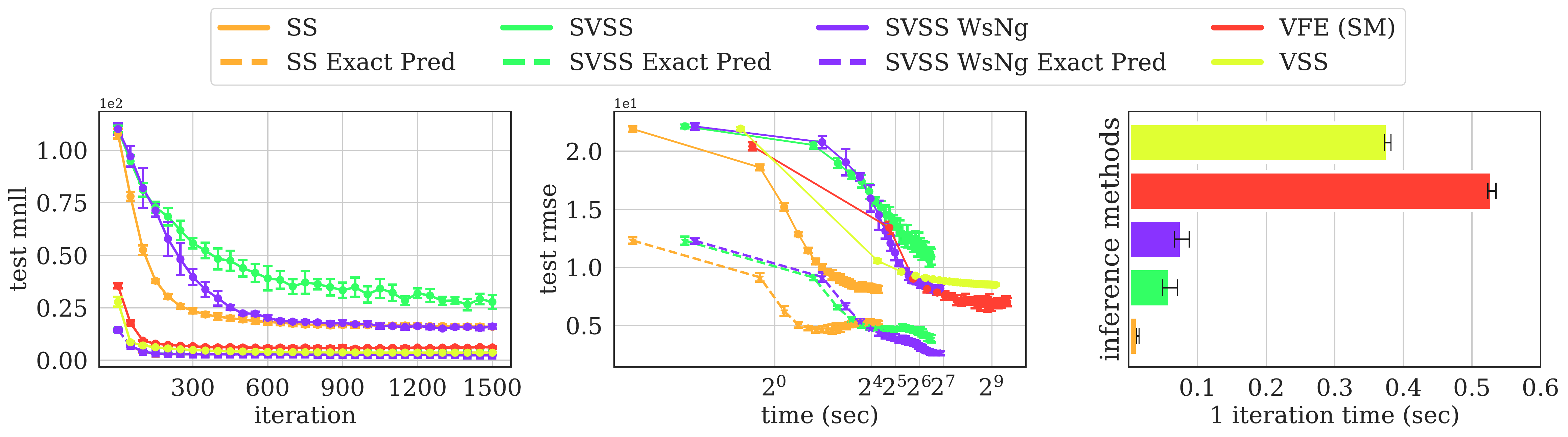}}
    \vspace{-1em}

\caption{Comparison of the inference methods for $\mathrm{CTslice}^{*}$ ($N=53,500 \ D=385$): Mean negative log likelihood (MNLL), RMSE, and single iteration time are used as performance metrics.}
\end{figure*}

%%%%%%%%%%%%%%%%%%%%%%%%%%%%%%%%%%%%%%%%%%%%%%%%%%%%%%%%%%%%%%%%%%%%%%%%%%%%%%%%%%%
%new table
%%%%%%%%%%%%%%%%%%%%%%%%%%%%%%%%%%%%%%%%%%%%%%%%%%%%%%%%%%%%%%%%%%%%%%%%%%%%%%%%%%%
\newcommand\Tstrut{\rule{0pt}{2.6ex}}         % = `top' strut
\newcommand\Bstrut{\rule[-0.9ex]{0pt}{0pt}}   % = `bottom' strut
\begin{table}[h]
  \caption{Regression task on large scale UCI datasets: We use the SM kernel ($Q=4$) and the number of total spectral points $M=4\times15$ for all inference methods. For VFE, we set the number of inducing points as $2M$ such that the size of VFE kernel matrix is equal to the other inference methods.}
  \label{sample-table}
  \centering
    \scriptsize
    \setlength\tabcolsep{1.25pt}
    \begin{tabular}{lrc ccccc ccc}
    \toprule
    &&& \multicolumn{8}{c}{RMSE} \\
    \cmidrule(r){4-11}
    Dataset     & $N$  & $d$ &    VFE (RBF) &  VFE (SM)  & SS  & VSS &
    \textbf{SVSS} & \textbf{SVSS Ws} & \textbf{SVSS WsNg} & \textbf{Exact (WsNg)}     \\
    % \cmidrule(r){1-8}
    % \cmidrule(r){9-10} 
    \cmidrule(r){1-11}
    Concrete & 1,030  & 8    &
    0.377        $\pm$  .005          &
    0.402        $\pm$  .009          &
    0.421   $\pm$      .041  & 
    0.618      $\pm$  .000            &     
    0.396   $\pm$      .033  & 
    0.347   $\pm$      .009  &      
    \textbf{0.341}   $\pm$      .009  & 
    0.833   $\pm$      .009 \\
            
    Skillcraft  & 3,325  & 18  &
    0.300        $\pm$ .013           &
    0.309        $\pm$ .016           &
    0.320        $\pm$ .008            &
    0.310        $\pm$ .015                 &
    0.296        $\pm$ .009            &
    0.297 $\pm$ .013            &
    \textbf{0.295} $\pm$ .012     &
    0.310   $\pm$ .007 \\
    
    Parkinsons & 5,875  & 20  & 
    1.386        $\pm$ .096           &
    1.670        $\pm$ .343          &
    3.084        $\pm$ .250            &
    6.618        $\pm$ .161            &
    2.821        $\pm$ .100            &
    1.505        $\pm$ .115            &
    1.635        $\pm$ .148      &
    \textbf{0.528}        $\pm$ .068        \\
    
    Kin8nm & 8,192  & 8  & 
    0.145        $\pm$ .005           &
    0.148        $\pm$ .010            &
    0.167        $\pm$ .006           &
    0.226        $\pm$ .045                 &    
    0.177        $\pm$ .009             &
    0.130        $\pm$ .005           &
    0.128        $\pm$ .008        & 
    \textbf{0.080}        $\pm$ .002        \\

    % Kin8nh & 8,192  & 8  & 
    %              $\pm$            &
    %              $\pm$            &
    %              $\pm$            &
    %              $\pm$            &
    %              $\pm$            &
    %              $\pm$            &
    %              $\pm$             \\

    % Pumadyn8nm & 8,192  & 8  & 
    %              $\pm$            &
    %              $\pm$            &
    %              $\pm$            &
    %              $\pm$            &
    %              $\pm$            &
    %              $\pm$            &
    %              $\pm$             \\

    %Pumadyn8nh & 8,192  & 8  &
    % 3.214        $\pm$ .060           &
    % \textbf{3.134}        $\pm$ .051            &
    % 3.176        $\pm$ .057               &
    % 4.517        $\pm$ .076                &                     
    % 3.150             $\pm$ .064               &
    % 3.145             $\pm$ .041               &
    % 3.139             $\pm$ .052               \\

    % Pumadyn-nm & 8,192  & 32  & 
    %         $\pm$            &
    %         $\pm$            &
    %         $\pm$            &
    %         $\pm$            &
    %         $\pm$            &
    %         $\pm$            &
    %         $\pm$             \\            

    Elevators & 16,599  & 18  &
    0.121   $\pm$      .003       &
    0.103        $\pm$ .002           &
    0.100        $\pm$ .005            &
    0.124        $\pm$ .003                &    
    0.093        $\pm$ .001           &
    0.095       $\pm$ .001           &
    0.095       $\pm$ .001        & 
    \textbf{0.089}        $\pm$ .002         \\

    \hline
    \Tstrut
    $\text{Protein}^{*}$ &  45,730  & 9  &
    0.613   $\pm$        .019  &    
    0.620    $\pm$  .019     & 
    0.631    $\pm$  .018     & 
    0.653    $\pm$  .016     & 
    0.621    $\pm$  .016     &
    0.603    $\pm$  .018     & 
    0.601    $\pm$  .016         & 
    \textbf{0.542}   $\pm$  .023       \\

    $\text{Blog}^{*}$ &  52,397  & 280  &
    0.915 $\pm$     .029    & 
    \textbf{0.771}    $\pm$ .015      & 
    0.885    $\pm$ .040      & 
    0.919    $\pm$ .067      &
    0.845    $\pm$ .013      &
    0.784    $\pm$ .034      & 
    0.784 $\pm$ .030      &
    0.846    $\pm$ .122            \\

    $\text{CTsilce}^{*}$ &  53,500  &  385  &
    6.522 $\pm$  .458        &    
    6.948 $\pm$  .329       &    
    8.122   $\pm$ .331      &       
    8.496  $\pm$  .106               &    
    10.988    $\pm$ .246      &       
    7.952    $\pm$ .169    &    
    7.867    $\pm$ .292 &
    \textbf{2.614}    $\pm$ .175     \\     
    \bottomrule
  \end{tabular}
\end{table}

After 5 repetitive experiments, the statistical results are obtained. For each experiment, the training and test data are randomly selected with a ratio of 9:1. We use single GPU (V100-16GB). For the Protein, Blog, and CTslice datasets, VFE (SM) incurs a memory problem. For a fair comparison, we equally divide 5 partitions of the dataset and then obtain the averaged result as regression task in \citep{wilson2016deep}.

Figure 5 compares the performance of the regression task for the CTslice datasets. We also present the prediction results obtained using the exact SM kernel with the parameters estimated by the SVSS and SVSS-WsNg. We can see that the parameters inferred by the proposed inference methods can be used for the exact SM kernel; The proposed inference method predicts the outputs more accurately using less computational time than other baseline inferences. Table 1 summarizes the RMSE for each dataset. Exact (WsNg) denotes the prediction results obtained by the exact SM kernel with the parameters estimated using SVSS-WsNg. We confirm that the proposed SVSS-WsNg and Exact (WsNg) achieve better prediction results for most of the dataset.

\section{Conclusion}
In this research, we proposed a way to efficiently estimate the hyperparameters of an SM kernel by employing a sampling-based variational inference. Because we employ a regularized ELBO estimator as an objective function, we can relax the over-fitting issue in SM kernel training . In addition, we train the parameters of SM kernel in a scalable manner for large-scale data. To improve the inference quality, we propose a sampling strategy for spectral points to robustly compute the regularized ELBO estimator. We also propose an approximate natural gradient to optimize the variational parameters of the SM kernel. We validated that the combination of the sampling strategy and the approximate natural gradient used in the proposed approximate inference accelerates the convergence of the parameters and results in better parameters for the SM kernel.

\section{Broader Impact}

In general, $\mathcal{GP}$ model is said to have the advantage of quantifying the uncertainty for the prediction of the model. This characteristic allows the  $\mathcal{GP}$ model to be widely used for the decision-making because the quantified uncertainty of the prediction can be incorporated into decision-making. In particular, for the sensitive problems where the individual randomness is more reflected in the dataset and makes decision difficult, the quantified uncertainty can be a helpful factor to the decision. For example, when the newly developed medicine should be verified for its use, the deterministic prediction of clinical effect for the potential users would not be completely trustful because the clinical trial results used as a training dataset contain the individual error of the tester. Thus, the credibility of the prediction could be important in determining whether developed medicine is used. This example explains why the uncertainty about the prediction should be estimated accurately and explains why the elaborated  $\mathcal{GP}$-based hybrid model have been proposed for the accurate uncertainty estimation and prediction. However, when the model becomes more complex, the learning of the model is likely to have problems as the case of the SM kernel. In this context, our approximate inference method can be used to train the complex $\mathcal{GP}$-based model using a large-scale dataset while alleviating the over-fitting issue. Furthermore, this approximate inference method can potentially help the process of making more reliable decisions.

%\newpage
\bibliography{neurips_citation}

\begin{thebibliography}{23}
\providecommand{\natexlab}[1]{#1}
\providecommand{\url}[1]{\texttt{#1}}
\expandafter\ifx\csname urlstyle\endcsname\relax
  \providecommand{\doi}[1]{doi: #1}\else
  \providecommand{\doi}{doi: \begingroup \urlstyle{rm}\Url}\fi

\bibitem[Bochner(1959)]{bochner1959lectures}
Salomon Bochner.
\newblock \emph{Lectures on Fourier integrals}.
\newblock Princeton University Press, 1959.

\bibitem[Wilson and Adams(2013)]{wilson2013gaussian}
Andrew Wilson and Ryan Adams.
\newblock Gaussian process kernels for pattern discovery and extrapolation.
\newblock In \emph{International Conference on Machine Learning}, pages
  1067--1075, 2013.

\bibitem[Kostantinos(2000)]{kostantinos2000gaussian}
N~Kostantinos.
\newblock Gaussian mixtures and their applications to signal processing.
\newblock \emph{Advanced signal processing handbook: theory and implementation
  for radar, sonar, and medical imaging real time systems}, pages 3--1, 2000.

\bibitem[Ulrich et~al.(2015)Ulrich, Carlson, Dzirasa, and Carin]{ulrich2015gp}
Kyle~R Ulrich, David~E Carlson, Kafui Dzirasa, and Lawrence Carin.
\newblock Gp kernels for cross-spectrum analysis.
\newblock In \emph{Advances in neural information processing systems}, pages
  1999--2007, 2015.

\bibitem[Parra and Tobar(2017)]{parra2017spectral}
Gabriel Parra and Felipe Tobar.
\newblock Spectral mixture kernels for multi-output gaussian processes.
\newblock In \emph{Advances in Neural Information Processing Systems}, pages
  6681--6690, 2017.

\bibitem[Remes et~al.(2017)Remes, Heinonen, and Kaski]{remes2017non}
Sami Remes, Markus Heinonen, and Samuel Kaski.
\newblock Non-stationary spectral kernels.
\newblock In \emph{Advances in Neural Information Processing Systems}, pages
  4642--4651, 2017.

\bibitem[Warnes and Ripley(1987)]{warnes1987problems}
JJ~Warnes and BD~Ripley.
\newblock Problems with likelihood estimation of covariance functions of
  spatial gaussian processes.
\newblock \emph{Biometrika}, 74\penalty0 (3):\penalty0 640--642, 1987.

\bibitem[Rasmussen(2004)]{rasmussen2004gaussian}
Carl~Edward Rasmussen.
\newblock Gaussian processes in machine learning.
\newblock In \emph{Advanced lectures on machine learning}, pages 63--71.
  Springer, 2004.

\bibitem[Kingma and Welling(2013)]{kingma2013auto}
Diederik~P Kingma and Max Welling.
\newblock Auto-encoding variational bayes.
\newblock \emph{arXiv preprint arXiv:1312.6114}, 2013.

\bibitem[Rahimi and Recht(2008)]{rahimi2008random}
Ali Rahimi and Benjamin Recht.
\newblock Random features for large-scale kernel machines.
\newblock In \emph{Advances in neural information processing systems}, pages
  1177--1184, 2008.

\bibitem[Lazaro-Gredilla et~al.(2010)Lazaro-Gredilla, Quinonero-Candela,
  Rasmussen, and Figueiras-Vidal]{lazaro2010sparse}
Miguel Lazaro-Gredilla, Joaquin Quinonero-Candela, Carl~Edward Rasmussen, and
  Anibal~R Figueiras-Vidal.
\newblock Sparse spectrum gaussian process regression.
\newblock \emph{Journal of Machine Learning Research}, 11:\penalty0 1865--1881,
  2010.

\bibitem[Amari(1998)]{amari1998natural}
Shun-Ichi Amari.
\newblock Natural gradient works efficiently in learning.
\newblock \emph{Neural computation}, 10\penalty0 (2):\penalty0 251--276, 1998.

\bibitem[Titsias(2009)]{titsias2009variational}
Michalis Titsias.
\newblock Variational learning of inducing variables in sparse gaussian
  processes.
\newblock In \emph{Artificial Intelligence and Statistics}, pages 567--574,
  2009.

\bibitem[Hensman et~al.(2013)Hensman, Fusi, and Lawrence]{hensman2013gaussian}
James Hensman, Nicolo Fusi, and Neil~D Lawrence.
\newblock Gaussian processes for big data.
\newblock In \emph{Uncertainty in Artificial Intelligence}, page 282. Citeseer,
  2013.

\bibitem[Hensman et~al.(2015)Hensman, Matthews, and
  Ghahramani]{hensman2015scalable}
James Hensman, Alexander~G Matthews, and Zoubin Ghahramani.
\newblock Scalable variational gaussian process classification.
\newblock \emph{Proceedings of Machine Learning Research}, 38:\penalty0
  351--360, 2015.

\bibitem[Gal and Turner(2015)]{gal2015improving}
Yarin Gal and Richard Turner.
\newblock Improving the gaussian process sparse spectrum approximation by
  representing uncertainty in frequency inputs.
\newblock In \emph{International Conference on Machine Learning}, pages
  655--664, 2015.

\bibitem[Hoang et~al.(2017)Hoang, Hoang, and Low]{hoang2017generalized}
Quang~Minh Hoang, Trong~Nghia Hoang, and Kian~Hsiang Low.
\newblock A generalized stochastic variational bayesian hyperparameter learning
  framework for sparse spectrum gaussian process regression.
\newblock In \emph{Thirty-First AAAI Conference on Artificial Intelligence},
  2017.

\bibitem[Dua and Graff(2017)]{Dua:2019}
Dheeru Dua and Casey Graff.
\newblock {UCI} machine learning repository, 2017.
\newblock URL \url{http://archive.ics.uci.edu/ml}.

\bibitem[Yang et~al.(2015)Yang, Wilson, Smola, and Song]{yang2015carte}
Zichao Yang, Andrew Wilson, Alex Smola, and Le~Song.
\newblock A la carte--learning fast kernels.
\newblock In \emph{Artificial Intelligence and Statistics}, pages 1098--1106,
  2015.

\bibitem[Le et~al.(2013)Le, Sarl{\'o}s, and Smola]{le2013fastfood}
Quoc Le, Tam{\'a}s Sarl{\'o}s, and Alex Smola.
\newblock Fastfood-approximating kernel expansions in loglinear time.
\newblock In \emph{Proceedings of the international conference on machine
  learning}, volume~85, 2013.

\bibitem[Jordan et~al.(1999)Jordan, Ghahramani, Jaakkola, and
  Saul]{jordan1999introduction}
Michael~I Jordan, Zoubin Ghahramani, Tommi~S Jaakkola, and Lawrence~K Saul.
\newblock An introduction to variational methods for graphical models.
\newblock \emph{Machine learning}, 37\penalty0 (2):\penalty0 183--233, 1999.

\bibitem[Hoffman et~al.(2013)Hoffman, Blei, Wang, and
  Paisley]{hoffman2013stochastic}
Matthew~D Hoffman, David~M Blei, Chong Wang, and John Paisley.
\newblock Stochastic variational inference.
\newblock \emph{The Journal of Machine Learning Research}, 14\penalty0
  (1):\penalty0 1303--1347, 2013.

\bibitem[Wilson et~al.(2016)Wilson, Hu, Salakhutdinov, and
  Xing]{wilson2016deep}
Andrew~Gordon Wilson, Zhiting Hu, Ruslan Salakhutdinov, and Eric~P Xing.
\newblock Deep kernel learning.
\newblock In \emph{Artificial Intelligence and Statistics}, pages 370--378,
  2016.

\end{thebibliography}

\end{document}